\PassOptionsToPackage{unicode}{hyperref}
\PassOptionsToPackage{hyphens}{url}
\PassOptionsToPackage{dvipsnames,svgnames,x11names}{xcolor}
\documentclass[
]{article}
\usepackage{amsmath,amssymb}
\usepackage{lmodern}
\usepackage{iftex}
\ifPDFTeX
  \usepackage[T1]{fontenc}
  \usepackage[utf8]{inputenc}
  \usepackage{textcomp} 
\else 
  \usepackage{unicode-math}
  \defaultfontfeatures{Scale=MatchLowercase}
  \defaultfontfeatures[\rmfamily]{Ligatures=TeX,Scale=1}
\fi
\IfFileExists{upquote.sty}{\usepackage{upquote}}{}
\IfFileExists{microtype.sty}{
  \usepackage[]{microtype}
  \UseMicrotypeSet[protrusion]{basicmath} 
}{}
\makeatletter
\@ifundefined{KOMAClassName}{
  \IfFileExists{parskip.sty}{%
    \usepackage{parskip}
  }{
    \setlength{\parindent}{0pt}
    \setlength{\parskip}{6pt plus 2pt minus 1pt}}
}{
  \KOMAoptions{parskip=half}}
\makeatother
\usepackage{xcolor}
\usepackage{color}
\usepackage{fancyvrb}

\DefineVerbatimEnvironment{Highlighting}{Verbatim}{commandchars=\\\{\}}
\newenvironment{Shaded}{}{}

\newcommand{\BuiltInTok}[1]{\textcolor[rgb]{0.00,0.50,0.00}{#1}}

\newcommand{\CommentTok}[1]{\textcolor[rgb]{0.38,0.63,0.69}{\textit{#1}}}

\newcommand{\ControlFlowTok}[1]{\textcolor[rgb]{0.00,0.44,0.13}{\textbf{#1}}}

\newcommand{\DecValTok}[1]{\textcolor[rgb]{0.25,0.63,0.44}{#1}}

\newcommand{\ImportTok}[1]{\textcolor[rgb]{0.00,0.50,0.00}{\textbf{#1}}}

\newcommand{\NormalTok}[1]{#1}
\newcommand{\OperatorTok}[1]{\textcolor[rgb]{0.40,0.40,0.40}{#1}}

\newcommand{\StringTok}[1]{\textcolor[rgb]{0.25,0.44,0.63}{#1}}
\newcommand{\VariableTok}[1]{\textcolor[rgb]{0.10,0.09,0.49}{#1}}

\usepackage{longtable,booktabs,array}
\usepackage{calc} 
\usepackage{etoolbox}
\makeatletter
\patchcmd\longtable{\par}{\if@noskipsec\mbox{}\fi\par}{}{}
\makeatother
\IfFileExists{footnotehyper.sty}{\usepackage{footnotehyper}}{\usepackage{footnote}}
\makesavenoteenv{longtable}
\usepackage{graphicx}
\makeatletter
\def\maxwidth{\ifdim\Gin@nat@width>\linewidth\linewidth\else\Gin@nat@width\fi}
\def\maxheight{\ifdim\Gin@nat@height>\textheight\textheight\else\Gin@nat@height\fi}
\makeatother
\setkeys{Gin}{width=\maxwidth,height=\maxheight,keepaspectratio}
\makeatletter
\def\fps@figure{htbp}
\makeatother
\providecommand{\tightlist}{%
  \setlength{\itemsep}{0pt}\setlength{\parskip}{0pt}}
\NewDocumentCommand\citeproctext{}{}
\NewDocumentCommand\citeproc{mm}{%
  \begingroup\def\citeproctext{#2}\cite{#1}\endgroup}
\makeatletter
 \let\@cite@ofmt\@firstofone
 \def\@biblabel#1{}
 \def\@cite#1#2{{#1\if@tempswa , #2\fi}}
\makeatother
\newlength{\cslhangindent}
\newlength{\csllabelwidth}
\newenvironment{CSLReferences}[2] 
 {\begin{list}{}{%
  \setlength{\itemindent}{0pt}
  \setlength{\leftmargin}{0pt}
  \setlength{\parsep}{0pt}
  \ifodd #1
   \setlength{\leftmargin}{\cslhangindent}
   \setlength{\itemindent}{-1\cslhangindent}
  \fi
  \setlength{\itemsep}{#2\baselineskip}}}
 {\end{list}}
\usepackage{calc}

\ifLuaTeX
\usepackage[bidi=basic]{babel}
\else
\usepackage[bidi=default]{babel}
\fi
\babelprovide[main,import]{american}

\def\languageshorthands#1{}
\ifLuaTeX
  \usepackage{selnolig}  
\fi
\IfFileExists{bookmark.sty}{\usepackage{bookmark}}{\usepackage{hyperref}}
\IfFileExists{xurl.sty}{\usepackage{xurl}}{} 
\hypersetup{
  pdftitle={Renard: A Modular Pipeline for Extracting Character Networks
from Narrative Texts},
  pdfauthor={Arthur Amalvy, Vincent Labatut, Richard Dufour},
  pdflang={en-US},
  colorlinks=true,
  linkcolor={Maroon},
  filecolor={Maroon},
  citecolor={Blue},
  urlcolor={Blue},
  pdfcreator={LaTeX via pandoc}}

\title{Renard: A Modular Pipeline for Extracting Character Networks from
Narrative Texts}

\usepackage[affil-it]{authblk}
\usepackage{orcidlink}
\author[1%
  ]{Arthur Amalvy%
    \,\orcidlink{0000-0003-4629-0923}\,%
    }
\author[1%
  ]{Vincent Labatut%
    \,\orcidlink{0000-0002-2619-2835}\,%
    }
\author[2%
  ]{Richard Dufour%
    \,\orcidlink{0000-0003-1203-9108}\,%
    }

\affil[1]{Laboratoire Informatique d'Avignon, France}
\affil[2]{Laboratoire des Sciences du Numérique de Nantes, France}
\date{29 February 2024}

\begin{document}
\maketitle

\section{Summary}\label{summary}

Renard (\emph{Relationships Extraction from NARrative Documents}) is a
Python library that allows users to define custom natural language
processing (NLP) pipelines to extract character networks from narrative
texts. Contrary to the few existing tools, Renard can extract
\emph{dynamic} networks, as well as the more common static networks.
Renard pipelines are modular: users can choose the implementation of
each NLP subtask needed to extract a character network. This allows
users to specialize pipelines to particular types of texts and to study
the impact of each subtask on the extracted network.

\section{Statement of Need}\label{statement-of-need}

Character networks (i.e., graphs where vertices represent characters and
edges represent their relationships) extracted from narrative texts are
useful in a number of applications, from visualization to literary
analysis (\citeproc{ref-labatut-2019}{Labatut \& Bost, 2019}). There are
different ways of modeling relationships (co-occurrences, conversations,
actions, etc.), and networks can be static or dynamic (i.e., series of
networks representing the evolution of relationships through time). This
variety means one can extract different kinds of networks depending on
the targeted applications. While some authors extract these networks by
relying on manually annotated data
(\citeproc{ref-park-2013-character_networks}{Park, Kim, \& Cho, 2013};
\citeproc{ref-park-2013-character_networksb}{Park, Kim, Hwang, et al.,
2013}; \citeproc{ref-rochat-2014-phd_thesis}{Rochat, 2014},
\citeproc{ref-rochat-2015-character_networks_zola}{2015};
\citeproc{ref-rochat-2017}{Rochat \& Triclot, 2017}), it is a
time-costly endeavor, and the fully automatic extraction of these
networks is therefore of interest. Unfortunately, there are only a few
existing software packages and tools that can extract character networks
(\citeproc{ref-metrailler-2023-charnetto}{Métrailler, 2023};
\citeproc{ref-sparavigna-2015-chaplin}{Sparavigna \& Marazzato, 2015}),
but none of these can output dynamic networks. Furthermore,
automatically extracting a character network requires solving several
successive natural language processing tasks, such as named entity
recognition (NER) or coreference resolution, and algorithms carrying
these tasks are bound to make errors. To our knowledge, the cascading
impact of these errors on the quality of the extracted networks has yet
to be studied extensively. This is an important issue since knowing
which tasks have more influence on the extracted networks would allow
prioritizing research efforts.

Renard is a fully configurable pipeline that can extract static and
dynamic networks from narrative texts. We base Renard on the generic
character network extraction framework highlighted by the survey of
Labatut \& Bost (\citeproc{ref-labatut-2019}{2019}). We design it so
that it is as modular as possible, which allows the user to select the
implementation of each extraction step as needed. This has several
advantages:

\begin{enumerate}
\def\labelenumi{\arabic{enumi}.}
\tightlist
\item
  Depending on the input text, the user can choose the most relevant
  series of steps and configure each of them as needed. Therefore, the
  pipeline can be specialized for different types of texts, allowing for
  better performance.
\item
  The pipeline can easily incorporate new advances in NLP, by simply
  implementing a new step when necessary.
\item
  One can study the impact of the performance of each step on the
  quality of the extracted networks.
\end{enumerate}

We intend for Renard to be used by digital humanities researchers as
well as NLP researchers and practitioners. The former category of users
can use Renard to quickly extract character networks for literary
analysis. Meanwhile, the latter can use Renard to easily represent
textual content using networks, which can be used as inputs for
downstream NLP tasks (classification, recommendation\ldots).

\section{Design and Main Features}\label{design-and-main-features}

Renard is centered about the concept of a \emph{pipeline}. In Renard, a
pipeline is a series of sequential \emph{steps} that are run one after
the other in order to extract a character network from a text. When
using Renard, the user simply \emph{describes} this pipeline in Python
by specifying this series of steps, and can apply it to different texts
afterwards. The following code block exemplifies that philosophy:

\begin{Shaded}
\begin{Highlighting}[]
\ImportTok{from}\NormalTok{ renard.pipeline }\ImportTok{import}\NormalTok{ Pipeline}
\ImportTok{from}\NormalTok{ renard.pipeline.tokenization }\ImportTok{import}\NormalTok{ NLTKTokenizer}
\ImportTok{from}\NormalTok{ renard.pipeline.ner }\ImportTok{import}\NormalTok{ NLTKNamedEntityRecognizer}
\ImportTok{from}\NormalTok{ renard.pipeline.character\_unification }\ImportTok{import}\NormalTok{ GraphRulesCharacterUnifier}
\ImportTok{from}\NormalTok{ renard.pipeline.graph\_extraction }\ImportTok{import}\NormalTok{ CoOccurrencesGraphExtractor}

\ControlFlowTok{with} \BuiltInTok{open}\NormalTok{(}\StringTok{"./my\_doc.txt"}\NormalTok{) }\ImportTok{as}\NormalTok{ f:}
\NormalTok{    text }\OperatorTok{=}\NormalTok{ f.read()}

\NormalTok{pipeline }\OperatorTok{=}\NormalTok{ Pipeline(}
\NormalTok{    [}
\NormalTok{        NLTKTokenizer(),}
\NormalTok{        NLTKNamedEntityRecognizer(),}
\NormalTok{        GraphRulesCharacterUnifier(min\_appearance}\OperatorTok{=}\DecValTok{10}\NormalTok{),}
        \CommentTok{\# users can pass \textquotesingle{}dynamic=True\textquotesingle{} and specify the}
        \CommentTok{\# \textquotesingle{}dynamic\_window\textquotesingle{} argument to extract a dynamic network}
        \CommentTok{\# instead of a static one.}
\NormalTok{        CoOccurrencesGraphExtractor(}
\NormalTok{            co\_occurrences\_dist}\OperatorTok{=}\DecValTok{10}\NormalTok{, dynamic}\OperatorTok{=}\VariableTok{False}
\NormalTok{        )}
\NormalTok{    ]}
\NormalTok{)}

\NormalTok{out }\OperatorTok{=}\NormalTok{ pipeline(text)}
\end{Highlighting}
\end{Shaded}

\begin{figure}
\centering
\includegraphics[width=\textwidth,height=0.3\textheight]{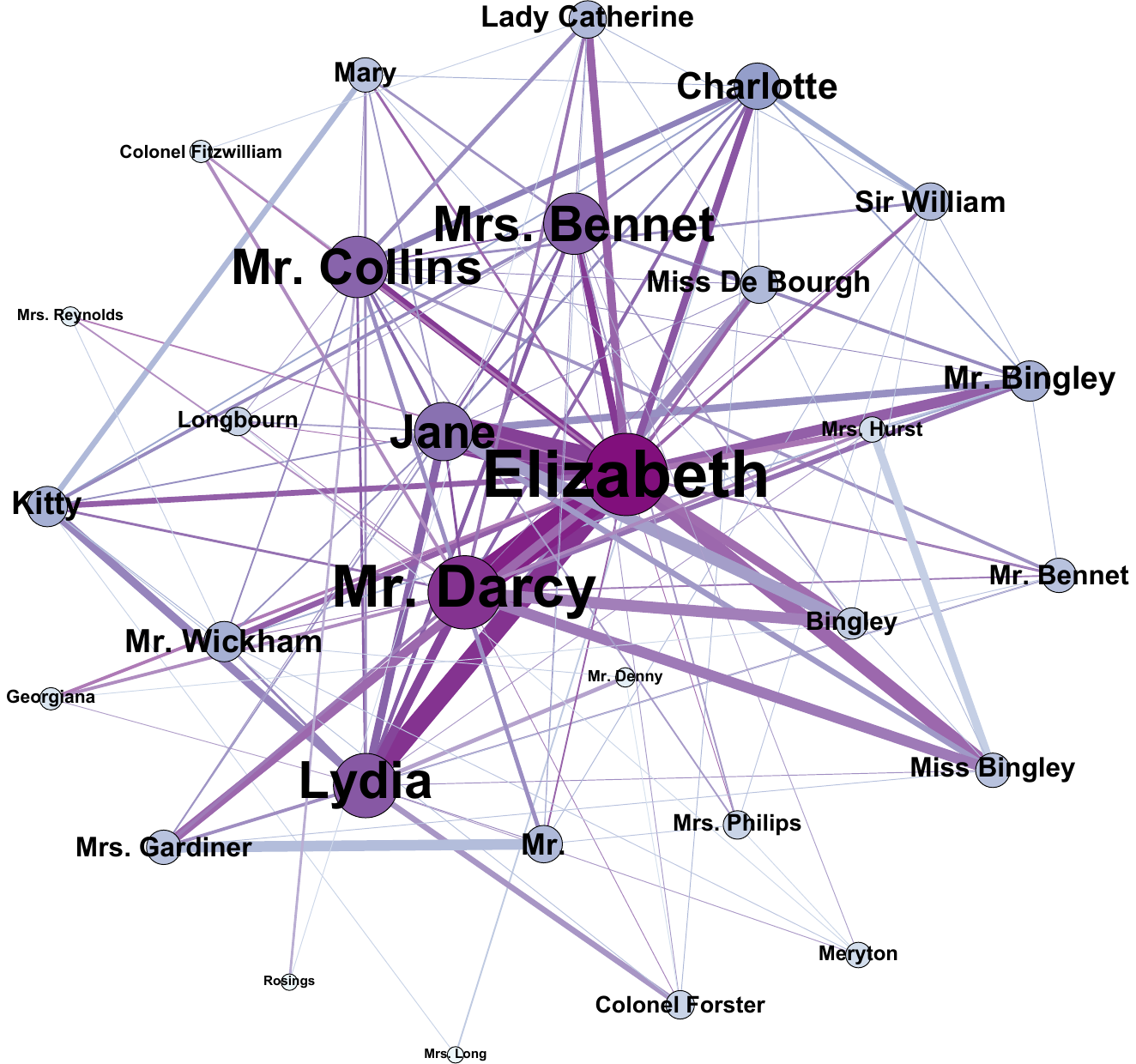}
\caption{Co-occurrence character network of Jane Austen's ``Pride and
Prejudice'', extracted automatically using Renard. Vertex size and color
denote degree, while edge thickness and color denote the number of
co-occurrences between two characters.}\label{fig:pp_network}
\end{figure}

As an example, Figure \ref{fig:pp_network} shows the co-occurrence
character network of Jane Austen's 1813 novel ``Pride and Prejudice'',
extracted using the Renard pipeline above. While this network is static,
users can also extract a dynamic network by passing the
\texttt{dynamic=True} argument to the last step of the pipeline, and
specifying the \texttt{dynamic\_window} argument: in that case, Renard
outputs a list of graphs corresponding to a dynamic network instead of a
single network\footnote{See
  \href{https://compnet.github.io/Renard/pipeline.html\#dynamic-graphs}{the
  documentation on dynamic networks} for more details.}. Renard uses the
NetworkX Python library (\citeproc{ref-hagberg-2008-networkx}{Hagbert et
al., 2008}) to manipulate graphs, ensuring compatibility with a wide
array of tools and formats.

To allow for custom needs, we design Renard to be very flexible. If a
step is not available in Renard, we encourage users to either:

\begin{itemize}
\tightlist
\item
  Externally perform the computation corresponding to the desired step,
  and inject the results back into the pipeline at runtime,
\item
  Implement their own step to integrate their custom processing into
  Renard by subclassing the existing \texttt{PipelineStep} class. If
  necessary, this \texttt{PipelineStep} can act as an adapter to an
  external process that may or may not be written in Python.
\end{itemize}

The flexibility of this approach introduces the possibility of creating
invalid pipelines because steps often require information computed by
previously run steps: for example, solving the NER task requires a
tokenized version of the input text. To counteract this issue, each step
therefore declares its requirements and the new information it produces,
which allows Renard to check whether a pipeline is valid, and to explain
at runtime to the user why it may not be\footnote{See
  \href{https://compnet.github.io/Renard/pipeline.html\#the-pipeline}{the
  documentation} for more details on steps requirements.}.

\begin{longtable}[]{@{}
  >{\raggedright\arraybackslash}p{(\columnwidth - 2\tabcolsep) * \real{0.6638}}
  >{\raggedright\arraybackslash}p{(\columnwidth - 2\tabcolsep) * \real{0.3362}}@{}}
\caption{Existing steps and their supported languages in Renard.
\label{tab:steps}}\tabularnewline
\toprule\noalign{}
\begin{minipage}[b]{\linewidth}\raggedright
Step
\end{minipage} & \begin{minipage}[b]{\linewidth}\raggedright
Supported Languages
\end{minipage} \\
\midrule\noalign{}
\endfirsthead
\toprule\noalign{}
\begin{minipage}[b]{\linewidth}\raggedright
Step
\end{minipage} & \begin{minipage}[b]{\linewidth}\raggedright
Supported Languages
\end{minipage} \\
\midrule\noalign{}
\endhead
\bottomrule\noalign{}
\endlastfoot
\texttt{StanfordCoreNLPPipeline} & eng \\
\texttt{CustomSubstitutionPreprocessor} & any \\
\texttt{NLTKTokenizer} & eng, fra, rus, ita, spa\ldots{} (12 other) \\
\texttt{QuoteDetector} & any \\
\texttt{NLTKNamedEntityRecognizer} & eng, rus \\
\texttt{BertNamedEntityRecognizer} & eng, fra \\
\texttt{BertCoreferenceResolver} & eng \\
\texttt{SpacyCorefereeCoreferenceResolver} & eng \\
\texttt{NaiveCharacterUnifier} & any \\
\texttt{GraphRulesCharacterUnifier} (inspired from Vala et al.
(\citeproc{ref-vala-2015-character_detection}{2015})) & eng, fra \\
\texttt{BertSpeakerDetector} & eng \\
\texttt{CoOccurencesGraphExtractor} & any \\
\texttt{ConversationalGraphExtractor} & any \\
\end{longtable}

Renard lets users select the targeted language of its their custom
pipelines. A pipeline can be configured to run in any language, as long
as each of its steps supports it. Table \ref{tab:steps} shows all the
currently available steps in Renard and their supported languages.

\section*{References}\label{references}
\addcontentsline{toc}{section}{References}

\phantomsection\label{refs}

\end{document}